# Title: A Soft Electronics-Free robot

Authors: E.-F. Markus Henke[1,*], Samuel Schlatter[2], Iain A. Anderson[1,*]

**Affiliations:**

[1] Auckland Bioengineering Institute, The University of Auckland, 70 Symonds Street, Auckland, New Zealand

[2] Microsystems for Space Technologies Laboratory, Ecole Polytechnique Fédérale de Lausanne, Lausanne, Switzerland

*Correspondence to: m.henke@auckland.ac.nz, i.anderson@auckland.ac.nz

**Abstract**: Biomimetic, entirely soft robots with animal-like behavior and integrated artificial nervous systems will open up totally new perspectives and applications. However, until now all presented studies on soft robots were limited to partly soft designs, since all designs at least needed conventional, stiff, electronics to sense, process signals and activate actuators. We present the first soft robot with integrated artificial nervous system entirely made of dielectric elastomers - and without any conventional stiff electronic parts. Supplied with only one external DC voltage, the robot autonomously generates all signals necessary to drive its actuators, and translates an in-plane electromechanical oscillation into a crawling locomotion movement. Thereby, all functional parts are made of polymer materials and carbon. Besides the basic design of the world's first entirely soft robot we present prospects to control general behavior of such robots.

**One Sentence Summary:** We present a novel class of soft, biomimetic robots based on multifunctional dielectric elastomers without electronic signal processing.

**Main Text:**

You never have to think about digestion. The enteric nervous system's continuous ganglionated nerves, located within the wall of the gut, respond to local muscle strain, switching living muscles on or off so that food is pushed forward along the digestive track, without control from the central nervous system (*1*). In this paper, we describe the emulation of the localized strain switching of living muscle using soft artificial muscle actuators with piezoresistive switches embedded within. Using a crawling robot, we demonstrate how switches and actuators can work alongside each other to produce an autonomous and lifelike metachronal movement without conventional control electronics. Such embedded control opens the door to a new range of soft robotic pumps, conveyors and crawling devices with integrated soft control that operate autonomously.

Presently, the vast majority of soft robots use pneumatic (*2-4*) or hydraulic (*5*) actuators. They can be made to operate in multi-freedom motion by using multiple chambers. A four-legged crawling robot by Shepherd et al. (*4*) can move by separate pneumatic actuation of each leg. In another instance an octopus-arm actuator can be steered using tendon-like cords (*6*). The bulk of the active elements are low modulus compliant materials like silicone or natural rubber, with Young's moduli on the order of $10^4$ Pa; very low compared with conventional robot motors, gears and fixtures built from materials with moduli in the range $10^9$-$10^{12}$ Pa (*7, 8*). But whether hard or soft they require external power and control units that are typically composed of dense and stiff materials. Thus, soft actuators can never be entirely soft in a complete assembly of actuator and controller. The integration of sensory functionalities into soft structures is even more challenging. Although there are promising developments in flexible and stretchable electronics (*9*), most conventional sensory technologies are not suited for integration into soft structures. To emulate the example of the gut, we need a technology that is soft and fully autonomous.

Within the range of smart materials, there is one with mechanical characteristics very similar to biological muscles (*10*): the dielectric elastomer (DE) (*11*). Its mechanism is electro-mechanical, offering the opportunity for rapid and fully integrated control of actuation. In its simplest embodiment a DE is a flexible capacitor consisting of a thin, pre-strained, material such as acrylic or silicone rubber. A membrane of typically 50 μm thickness is flanked on both sides by a stretchable electrode (Fig. S1). Application of charge to the electrodes produces an electrostatic Maxwell stress, resulting in in-plane expansion and across-thickness contraction. With impressive operating capabilities that include maximum areal strains in excess of 1692% (*12*), DE devices can be made to match and exceed the performance of natural muscles, earning the moniker of artificial muscles (*13*). In addition to being used as actuators with large actuation (*10, 12, 14*), they can perform as sensors for large strain (*15, 18*) and as power generators (*19*). Some of these functions can be performed simultaneously (*11*). For instance, dielectric elastomer actuators (DEAs) made touch sensitive through electronic capacitive sensing, can be placed in arrays so that they autonomously push an object forward, mimicking the mechanosensitive actuation of cilia (*21*).

The muscle activity of the gut, cited above, provides a clue on how to avoid bulky and heavy external controllers in a soft actuator: by using strain directly as an electronic switching mechanism. One embodiment would be a stretchable conductor integrated directly with a DE artificial muscle that can undergo large changes in resistivity with artificial muscle strain and, through this, turn electric charge flow on or off. Beruto et al. demonstrated that stretchable materials, such as silicon-graphite mixtures, can show large changes in piezoresistivity (in the order of $10^5$) at low voltages around the percolation threshold, useful for strain and pressure sensing (*22*). A switch that would be effective for DE muscle control would need to operate within the kilovolt range of DE.

O'Brien et al. (*23*) have identified a suitable high-voltage switching material, consisting of carbon particles in a grease, that can be imprinted on the surface of a DE membrane: the dielectric elastomer switch (DES). The conductivity of a DES can then be changed by several orders of magnitude by a voltage-induced elongation of an adjacent DEA (*23*). DEAs and DESs can be combined together as strain-dependent electrical signal inverters. When an odd number of these inverters is assembled in a closed loop cycle, one obtains a soft electro-mechanical oscillator that is entirely and uniquely controlled by elastomer strain (*24*), without any conventional electronic components being used.

To demonstrate the potential of DEs and DESs for soft, electronics-free, self-controlled robots, we have designed a robot with actuators, signal processing and circuitry composed solely of elastomer and printed carbon electrode (Fig. 1A). To achieve forward locomotion our requirement was to ensure that the DE elements actuated on and off in a self-regulating cycle. Therefore, the robot is comprised of an electronic network of three imprinted carbon grease resistors, three imprinted DESs and six DEAs that are able to generate an electromechanical, oscillating, in-plane motion ($x_1 x_2$-plane). The DE membrane is equibiaxially pre-strained by 300% and glued onto a rigid acrylic frame. The in-plane motion generated by the DEAs is translated into a forward motion by five V-shaped compliant legs, laser cut from 127μm thick PET film. Thus, each leg connects two adjacent DEAs (Fig. 1B). If a DEA is activated it elongates; this causes the leg beneath it to move up and forward (Fig. 1C). After the DEA's full elongation the DES network discharges this DEA and, at the same time, charges the one behind it (Fig. 1D). The leg moves down into contact with the supporting surface while the leg behind it moves up and forward (Fig. 1E). The self-generated signal travels from actuator to actuator from the front to the back of the robot producing an autonomous crawling motion.

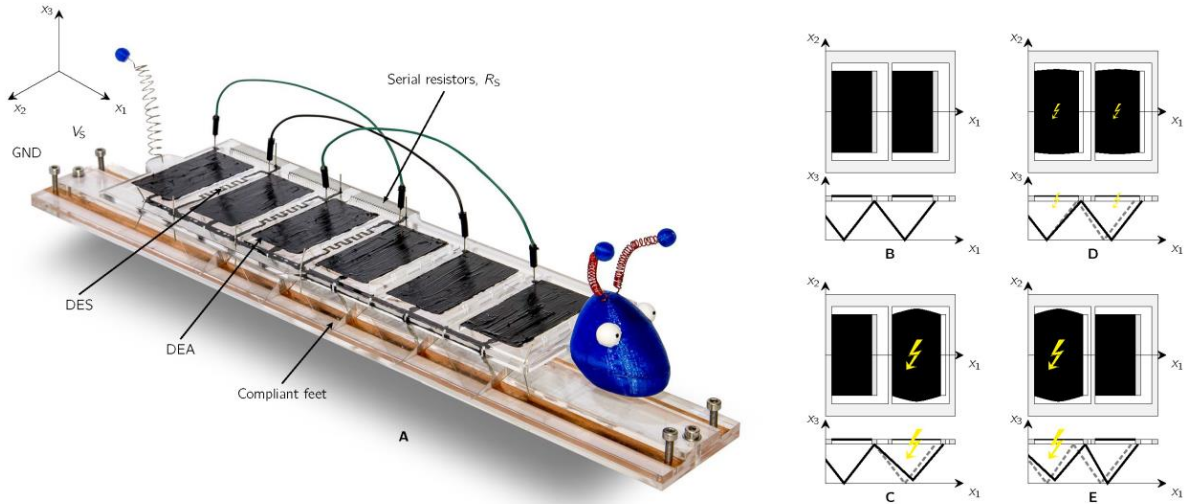

**Fig. 1 Robot design and principle of operation. (A)** Photograph of the demonstrator robot – Trevor the Caterpillar – containing no conventional electronic parts. All functional elements consist of DE structures (DEAs and DES) and imprinted carbon resistors. **(B)** Model of the robot's movement. The in-plane electromechanical traveling wave generated within the robot's DE network is translated into a forward motion by its flexible legs, which simultaneously connect the robot to its driving voltage and the surface.

To generate the necessary three high-voltage driving signals, the first three DEAs are combined with DESs and carbon grease resistors (Fig. 2A, B) at the top of the robot. The combination of DEA and DES represents a basic high-voltage logic unit: the digital inverter (Fig. 2D). In its initial state the switch is not conducting and has a resistance $R_{DES}>10^{12}\,\Omega$. If a supply voltage $V_S$ of typically 3000 V is applied to the DEA, it elongates and compresses the DES. That causes a significant resistance drop ($R_{DES}\approx1\dots5$ M$\Omega$) in the DES, resulting in a drop in the output voltage.

The resistance changes occur steeply, suitable for high-voltage switching (Fig. 2E). The three digital inverters are combined in a closed loop by feeding back the output signal of the last inverter to the first input (Fig. 2C). Combining an odd number of inverters in such a loop invokes a spontaneously activated oscillation signal between the high- and the low-voltage level (*24*) (Fig. S4). The generated voltage signals are the superposition of different wave forms, based on several, simultaneous, physical processes (Section S4) and are 120° phase shifted with respect to each other (Fig. 2C right). To enable a stable crawling motion there is an additional DEA in parallel with each inverter. In fact, the robot only requires a single DC input voltage and ground potential at the copper rails depicted in Fig. 2A, to which it is connected by conductive paint tracks on its feet. All signal processing and transforming of the in-plane movement of the DEAs into a crawling locomotion is entirely done by elastomer dielectric components and biomimetic compliant structures without any stiff common electronics.

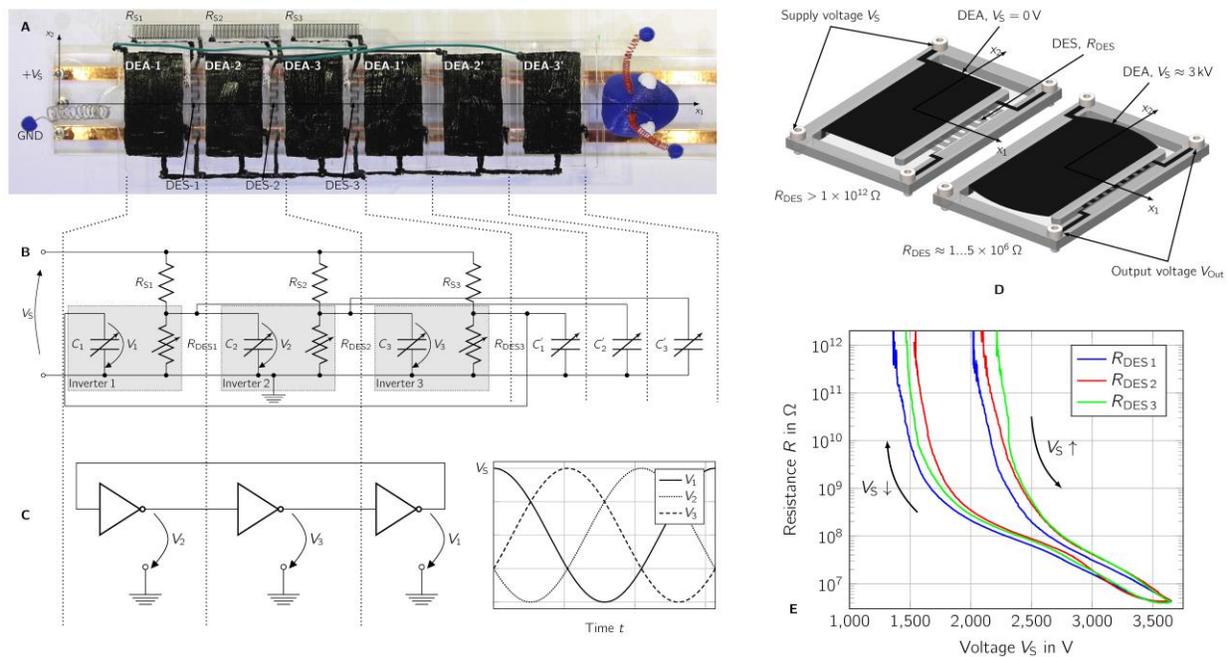

**Fig. 2 (A)** Photograph of the robot from above, depicting all dielectric electronic parts. **(B)** Electric model of the in-plane dielectric circuitry, emphasizing the three dielectric inverters. **(C)** basic model of a closed loop inverter cycle, which generates a high-voltage oscillation with three 120° phase shifted sinusoidal voltages $V_i$ (right). **(D)** Computer aided design model of a single dielectric inverter unit. Without a supply voltage $V_S$ the resistance $R_{DES}$ of the switch is $>10^{12}$ Ω (left). When a supply voltage $V_S \approx 3$ kV is applied the DEA expands, compresses the DES and $R_{DES}$ drops to $\approx 10^6$ Ω. **(E)** Resistance $R_{DES}$ of three example DES over the applied supply voltage $V_S$.

This electronics-free robot is able to fulfill a basic task – crawling along a rail – without any external control or conventional, internal signal processing units. Like the example for the gut, the movement of each DE artificial muscle is controlled by the mechanical strain of its neighbor, with strain coupled DES providing a "nervous system" that communicates both signal and excitatory charge for actuation, in a self-regulating system. This example of strain based actuation and control heralds a new class of entirely soft, autonomous, robots that produce self-controlling wave-like movement without additional electronics, and are composed entirely of low modulus materials.

Fig. 3 demonstrates the autonomous movement capabilities of our robot during several experiments. The self-generated electromechanical wave travelling through the robot's DE muscles, and compressing the corresponding DESs, is depicted in Fig. 3A-C. The resulting oscillation voltage signals, are shown in Fig. 3D together with the supplied DC voltage $V_S$. The signals show the expected 120° phase shift but differ in amplitude. This is caused by the varying switching voltages of the switches. The oscillation frequency can be controlled by the value of the serial resistors $R_S$ and the value of the supply voltage $V_S$ (Fig. 3E). This discovery enables the design of the robot's DE network so that it is specific for a desired crawling speed interval, with fine-tuning by adjusting the supply voltage $V_S$.

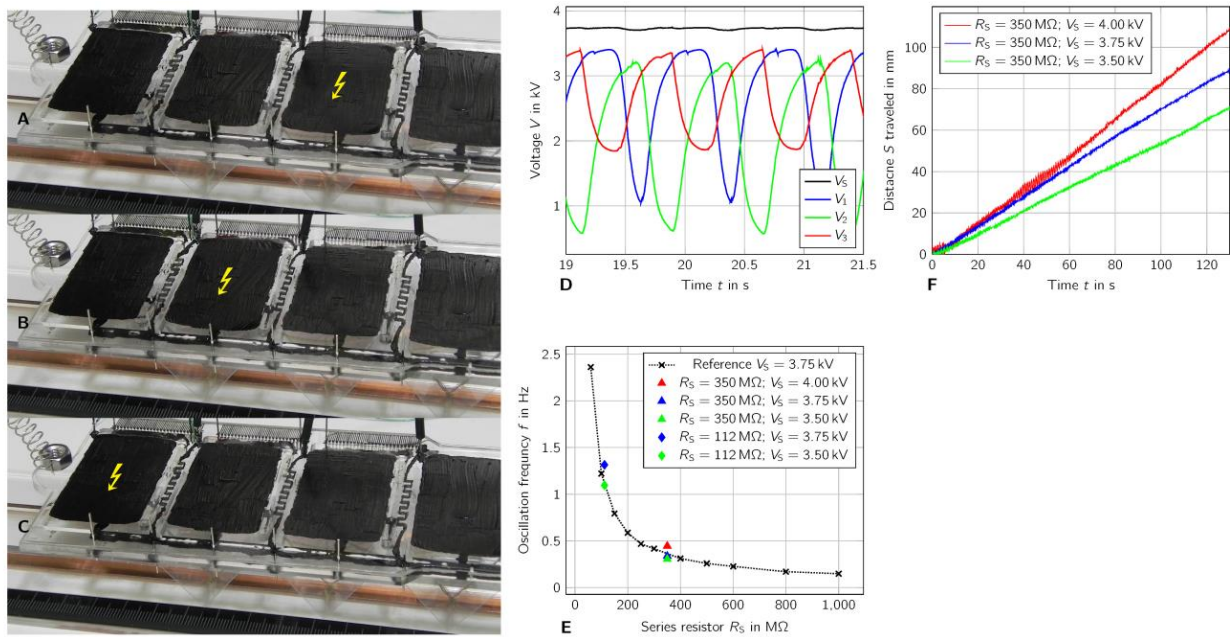

**Fig. 3 Experimental results for the electronic-free soft robot. (A-C)** Traveling high-voltage waveform along the robot's DE muscles and resulting compression of the corresponding DESs - flash symbols mark position of the highest voltage at each time period. **(D)** Measured oscillating voltages signals according to the nomenclature of (Fig. 2B/C). **(E)** Influence of serial resistor values $R_{Si}$ and the supply voltage $V_S$ on the oscillation frequency of the electromechanical wave. **(F)** Measured position for the robot during experiment for different supply voltages $V_S$.

The results show that it is possible to design biomimetic robots with actuators and charge control devices that are composed entirely of the same soft materials and without conventional, stiff, electronic components. There is a vast number of promising studies focusing on the development of soft and smart structures and impetus concepts (e.g. *25*). Our investigations accelerate the development of fully soft, animal-like, robots without any rigid materials. All used materials are suitable for modern production technologies such as 3D-, screen-, or ink jet-printing.

### Acknowledgement:

This work was supported by a fellowship within the Postdoc-Program of the German Academic Exchange Service (DAAD). We wish to thank Benjamin M. O'Brien for the concept of the oscillator-driven crawling robot and the helpful discussions we have had with him during the development of the final electronics-free prototype.

### Supplementary Materials:

Materials and Methods

Supplementary Text

Figures S1-S4

Movies S1-S2

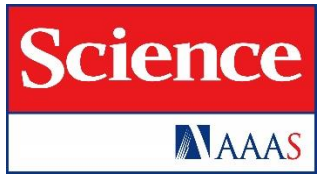

# Supplementary Materials for

## A Soft Electronic-Free Robot

E.-F. Markus Henke, Samuel Schlatter, Iain A. Anderson

correspondence to: m.henke@auckland.ac.nz, i.anderson@auckland.ac.nz

**This PDF file includes:**

    Materials and Methods
    Supplementary Text
    Figs. S1 to S4

    Captions for Movies S1 to S2

**Other Supplementary Materials for this manuscript includes the following:**

    Movies S1 to S2

**Materials and Methods**

Production of DEA and DES

The Dielectric Elastomer Actuators (DEAs) and Dielectric Elastomer Switches (DESs) used in the electronic-free robot are applied to a pre-strained VHB 4905 (by 3M) membrane. To ensure proper functionality of the actuators and the switches, either components require different, and specific, membrane pre-strains and, therefore, are applied in individual steps (Fig S1):

1. The frame for the robot is laser cut using a Trotec Speedy 300; simultaneously the locations for the carbon grease resistors are engraved
2. Nyogel 756G carbon grease is applied to the resistor tracks using a brush. The surplus grease is removed with a wipe.
3. At first, a piece of VHB 4905 is applied in an iris-like pre-strainer, which is set to an inner diameter of 85 mm.
4. The membrane is strained to a diameter of 147 mm
5. At that pre-strain the switching tracks are applied to the membrane by an imprint process:
    a. The switching grease, consisting of a 5:1 mixture by weight of Molykote 44 Medium grease to Cabot Vulcan XC72 carbon, is applied to a PDMS stamp (Sylgard 184, by Dow Corning) using a size zero soft taper point color shaper from Royal Sovereign Ltd. UK;
    b. The coated stamp is pressed on the top of the membrane, against a flat counter mold from the bottom of the membrane, for at least 30 seconds to ensure a proper transfer of the material;
    c. The above procedure is repeated for all necessary switches.
6. The membrane is further strained to a diameter of 294 mm
7. The robot frame is applied to the membrane from the bottom. At this point it is necessary that the switches are exactly positioned.
8. The connection bars which connect the DEAs to the feet are applied from the bottom.
9. The electrodes for the DEAs are hand brushed to the membrane using Nyogel 756G carbon grease.
10. All necessary conductive tracks are hand brushed onto the top and the bottom of the robot.
11. After finishing the circuitry, the completed robot is cut from the pre-strained membrane using a scalpel and the feet are applied. A proper connection of the conductive paths of the feet to the ground and high-voltage bus at the bottom of the robot is ensured by application of additional carbon black.
12. The robot is placed on the voltage supply tracks and starts crawling as soon as suitable supply voltage is applied.

Experimental Setup:

All experimental investigations described in this contribution were performed in the Biomimetics Lab at the Auckland Bioengineering Institute. As voltage supply, a Biomimetics EAP controller was used, remotely controlled by LabView software. Voltage measurements were performed with a self-developed, high-resistance voltage divider combined with a pre-amplifier to decrease the high voltages to low voltage signals, which were measured by a National Instruments SCB-68 analog to digital converter unit and processed by LabView. The position of the robot and, thus, its motion were measured by a laser distance sensor LDS70-200 by Eltrotec and also recorded and processed by LabView. Photographs were taken with a Canon EOS 70D digital single-lens reflex camera.

**Supplementary Text**

Dielectric Elastomer Actuators

DEAs (Fig. S2) consist of (usually) several 10µm thick, polymeric membranes, such as silicone or polyacryl, coated with compliant electrodes at top and bottom. To generate significant planar actuation, the membrane has to be pre-strained by an external stess $\sigma_{\text{ext } i}$ in both $x_1$- and $x_2$-direction. Typical pre-strains are $\lambda_{\text{pre}}$=1.3 … 3.5 and depend on the membrane material and the desired application. The pre-strain and the hyper-viscoelastic material behavior generate a three dimensional strain configuration in equilibrium. By applying a voltage of (typically) several thousand volts, a Maxwell pressure over the membrane thickness is caused by the attraction between opposite charges on the membrane. Since polymers are generally incompressible, the additional strain in $x_3$-direction causes the membrane to elongate in area along $x_1$ and $x_2$.

Dielectric Signal Inverters Based on DES

DESs are able to switch resistance over a wide range - up to six orders of magnitude. This makes them applicable to high-voltage logic gates. Our investigations in this field are continuing. The fundamental logic signal processing unit is the digital inverter, which was also used in the signal processing unit in the robot as described. Fig. S3 gives an overview of the design and functionality of a digital dielectric inverter unit. In Fig. S3A there is no input voltage $V_{\text{in}}$ applied to the inverter. The DEA is not actuated, the DES is strained over its percolation threshold and possesses a resistance $R_{\text{DES}}>10^{12}$ Ω. Thus, all the supply voltage drops over the resistance $R_{\text{DES}}$ and the output voltage is high. By applying a high (Fig. S3B) signal, usually the supply voltage, to the input, the DEA can be elongated. This causes the DES to contract. As a result, the resistance drops dramatically to typically several MΩ, pulling the output voltage to a low level. All voltages beneath the switching voltage of the DES – typically 1500 V – are considered as low signal level.

As shown in Fig. S3C, the output voltage is always the opposite of the input voltage. However, due to the remaining resistance value of $R_{\text{DES}}>1$ MΩ, the output voltage will never drop to 0 V, and the switching behavior is slow because of the viscoelastic properties of the membrane material used. Fig. S3D depicts the static elongation of the DEA and the compression of the switch at different input voltage values.

Detailed Description of the Signal Generation

Fig. S4 depicts an idealized electric model of the robot during an entire oscillation cycle with periodic time *T*. After applying a supply voltage $V_S$ to the oscillator, a self-primed oscillation, with three phase shifted voltages dropping over the individual inverters, is generated. The phase shift between the voltage signals is 120° $\left(\frac{2\pi}{3}\right)$. After a settling time the voltage signals oscillate with constant amplitude and frequency. The maximum voltage on a fully charged capacitor travels from DEA-3 to the left to DEA-1 and starts again from DEA-4. In Fig. S4A the capacitors representing DEA-3 and 3' are fully charged and the maximum voltage drops occur there. That causes the maximum elongation at DEA-3 and compresses DES-4. This switch then acts as a conductor and discharges the capacitors DEA-1 and 1'. Those actuators shrink to their initial state and switch DES-1 to high resistance, in effect an insulator compared to the other resistances. The absence of a conduction path along DES-1 forces the current through $R_{S1}$ to charge the discharged capacitor DEA-2. The increasing charge on DEA-2 causes that actuator to elongate and simultaneously lower the resistance of DES-2, which causes the capacitors DEA-3 and 3' to discharge via DES-2. After $T/3$ (Fig. S4B) the voltage on DEA-2 and 2' reaches its maximum value. Therefore, DES-2 is conducting and discharges capacitors DEA-3 and 3'. The switch DES-3 is in its high resistance state now and causes the current through $R_{S3}$ to charge capacitors DEA-1 and 1'. After $2T/3$ DEA-1 and 1' are fully charged, the switch DES-1 is conducting and discharges DEA-2 and 2' and DEA-3 starts charging again (Fig S4C) - restarting the overall cycle. As result, a continuous, self-primed, electro-mechanical, travelling wave from right to left is generated.

The frequency of the oscillation depends on the propagation delay $t_p$ of the individual inverters, which is mainly caused by the viscoelastic actuation of the VHB membrane.

Every DEA is connected to an end of one of the five feet of the robot. The left drawings in Fig. S4 show, in addition to the top view, also a schematic side view with the feet. Due to friction one foot only will move to the right, if the right upper connection to one of its DEAs moves to the right, and slightly out of the membrane plane, due to a voltage induced loss of membrane tension. The right adjacent foot, whose upper left end also moves to the right, won't take a step to the right because of its friction with the surface. That assumes a certain roughness of the surface on which the robot walks. However, this also ensures that the robot will always walk in the desired direction, although the electromechanical wave travels in the opposite direction.

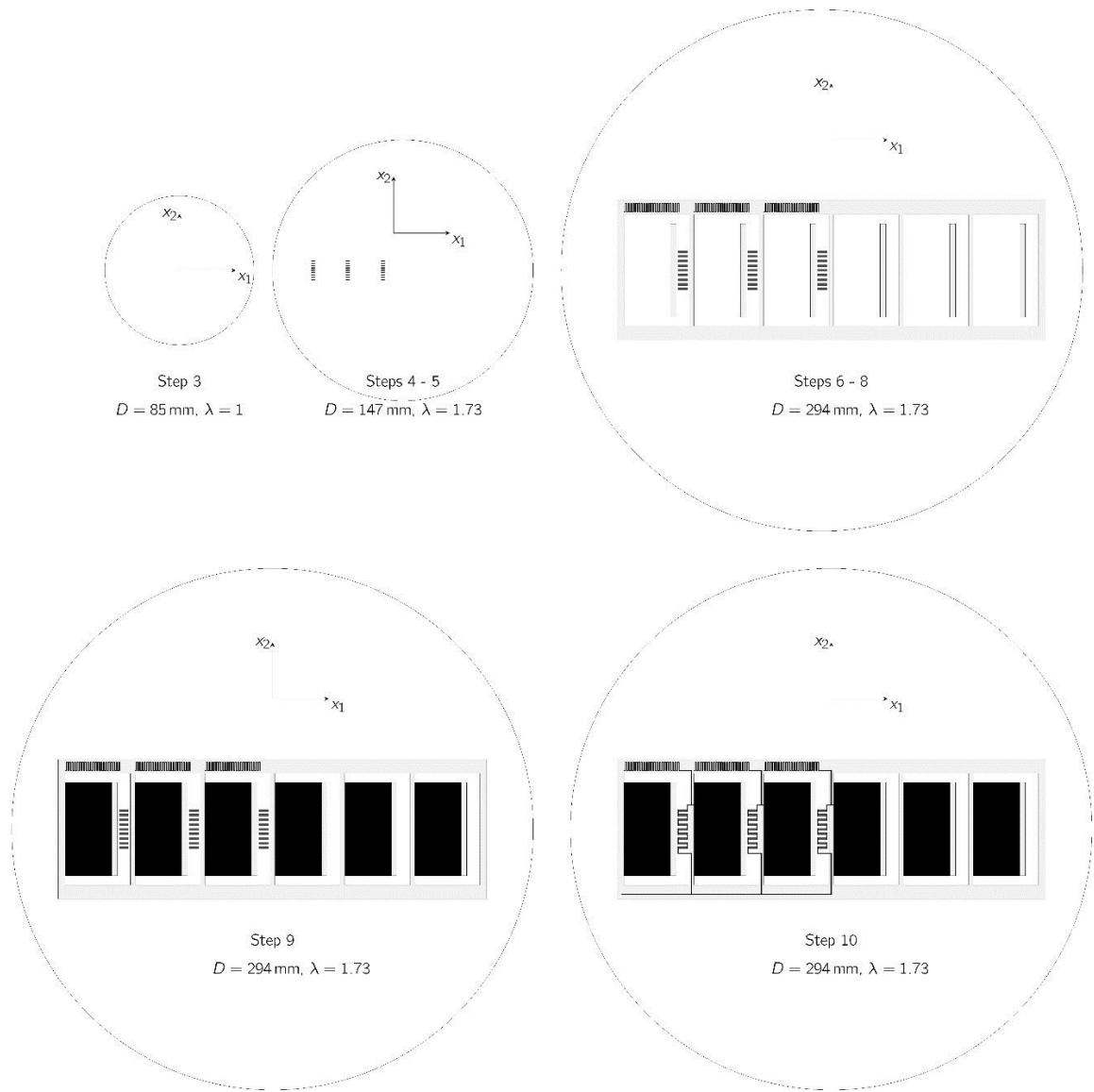

**Fig. S1**

Production steps to build up the presented electronic free robot, according to the description in the Materials and Methods section.

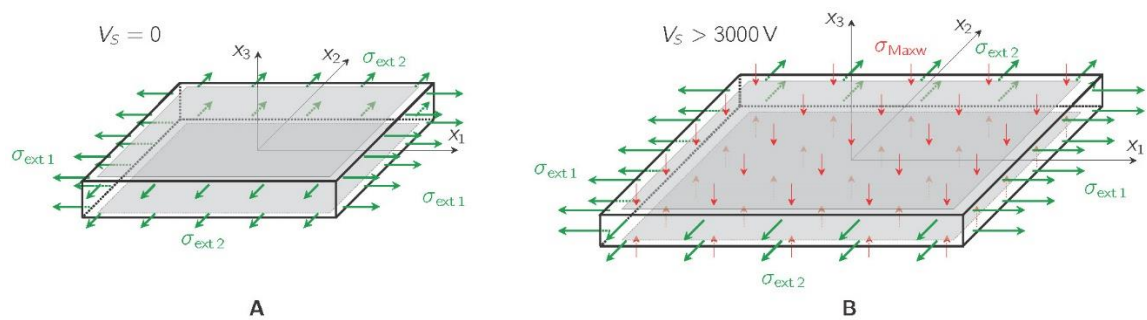

**Fig. S2**

Principle of a Dielectric Elastomer Actuator, consisting of an elastomer membrane coated with compliant electrodes on the top and the bottom. (A) The dielectric elastomer membrane is equibiaxially pre-strained by the external stresses $\sigma_{ext\,i}$. There is no supply voltage $V_S$ applied to the compliant electrodes. (B) A suitable supply voltage $V_S$ (typically > 3000 V) is applied to the compliant electrodes. That causes a Maxwell pressure along the thickness of the membrane. According to the governing equations the DEA will be compressed in x3-direction and elongate in direction $x_1$ and $x_2$. Combining such an DEA with a flexible mechanical structure makes an actuator.

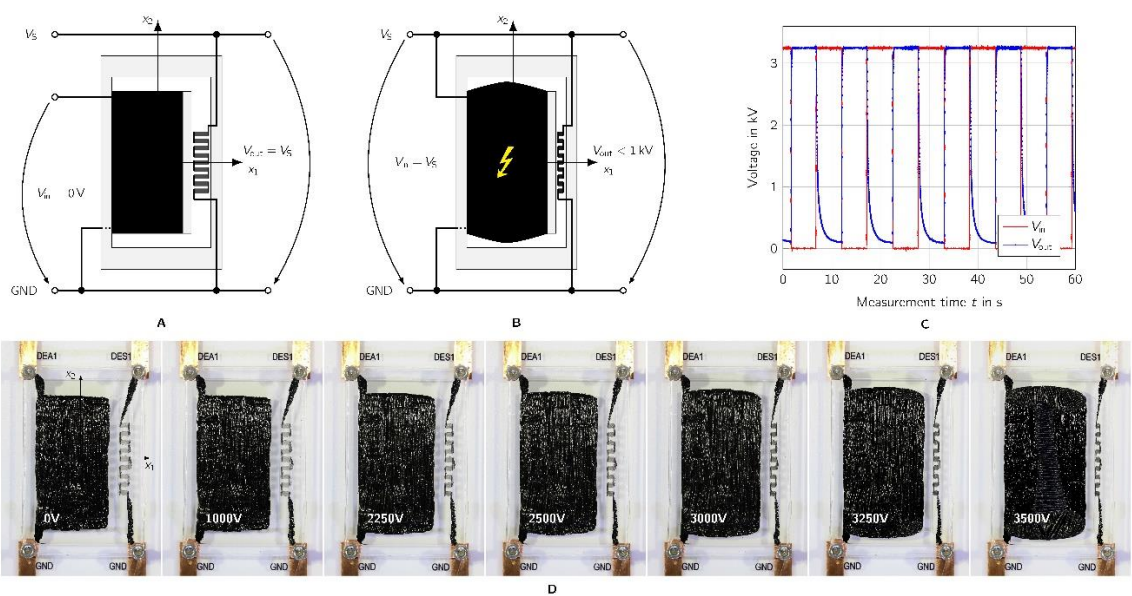

**Fig. S3**
Dielectric Elastomer high-voltage signal inverter. (A) In its initial state, when no input voltage Vin is supplied, the switch is pre-strained over its percolation threshold and does not conduct. The output voltage $V_{out}$ equals the supply voltage $V_S$. (B) If the input voltage $V_{in}=V_S$, the EAP elongates and compresses the switch. Due to its piezoresistive characteristics the resistance of the switch drops and so does the output voltage. (C) Output voltage $V_{out}$ for a square wave input voltage signal Vin over the measurement time. (D) Example pictures for DEA elongation and DES compression upon different time constant input voltages.

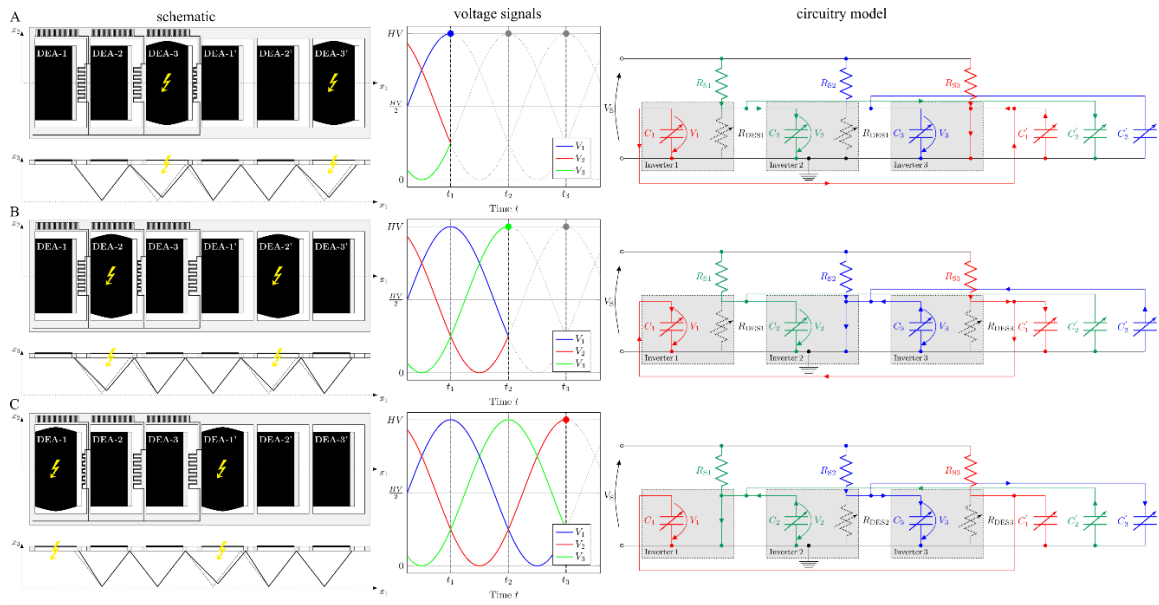

**Fig. S4**
Idealized signal generation in the robot at different times $t_i$ during one oscillation cycle. (Left) schematic, idealized elongation of DEAs, (middle) idealized voltage signals at moment $t_i$, (right) actual electric circuit condition: (A) moment $t_1$, DEA 1 and 1' discharging, DEA 2 and 2' charging, DEA 3 completely charged, DES 1 and DES 2 not conducting, DES 3 conducting; (B) moment $t_2$, DEA 1 and 1' charging, DEA 2 and 2' completely charged, DEA 3 discharging, DES 1 not conducting, DES 2 conducting, DES 3 not conducting; (C) moment $t_3$, DEA 1 and 1' completely charged, DEA 2 and 2' discharging, DEA 3 charging, DES 1 conducting, DES 2 and DES 3 not conducting.

**Movie S1**

This Movie shows the movement of the actual version of the robot without any conventional electronics, emphasizing the signal generation unit and the movement of the feet.

**Movie S2**

The movie shows and explains the designed robot in an earlier version, still containing conventional resistors. However, it gives an impression of the biologically-inspired movement generated by the dielectric elastomer actuator and switch network. The movie was produced for promotional purposes related to the Dielectric Elastomer Switch technology.